\PassOptionsToPackage{table}{xcolor}
\documentclass{article} 
\usepackage{mathpazo} 
\usepackage[final]{colm2025_conference}

\usepackage{microtype}
\usepackage{hyperref}
\usepackage{url}
\usepackage{booktabs}
\usepackage{xspace}
\usepackage{lineno}
\usepackage{wrapfig}
\usepackage{adjustbox}
\usepackage{multicol}
\usepackage{multirow}
\usepackage{ulem}
\useunder{\uline}{\ul}{}
\usepackage{subcaption}
\usepackage{scalerel}   
\usepackage{bbm}

\usepackage[table]{xcolor}  

\usepackage{amsmath,amsfonts,bm}

\def\eqref#1{equation~\ref{#1}}

\def\1{\bm{1}}

\DeclareMathAlphabet{\mathsfit}{\encodingdefault}{\sfdefault}{m}{sl}
\SetMathAlphabet{\mathsfit}{bold}{\encodingdefault}{\sfdefault}{bx}{n}

\usepackage{cleveref}
\definecolor{darkblue}{rgb}{0, 0, 0.5}
\hypersetup{colorlinks=true, citecolor=darkblue, linkcolor=darkblue, urlcolor=darkblue}

\definecolor{myblue}{RGB}{190, 242, 255}
\definecolor{myyellow}{RGB}{255, 244, 190}
\definecolor{myred}{RGB}{255, 225, 238}
\definecolor{mypurple}{RGB}{233, 225, 255}

\usepackage{graphicx}

\title{How does Watermarking Affect Visual Language Models in Document Understanding?}

\author{Chunxue Xu$^{\dagger}$,\quad
Yiwei Wang$^{\mathsection}$, \quad
Bryan Hooi$^{\ddagger}$, \quad
Yujun Cai$^{\diamond}$, \quad
Songze Li$^{\dagger}$\thanks{Corresponding Author.} 
\\
$^{\dagger}$ Southeast University, China 
\quad $^{\mathsection}$ University of California, Merced, USA \\
$^{\ddagger}$ National University of Singapore, Singapore \\
$^{\diamond}$ The University of Queensland, Australia
}

\begin{document}

\ifcolmsubmission
\linenumbers
\fi

\maketitle

\begin{abstract}

Visual Language Models (\vlm) have become foundational models for document understanding tasks, widely used in the processing of complex multimodal documents across domains such as finance, law, and academia. However, documents often contain noise-like information, such as watermarks, which inevitably leads us to inquire: \emph{Do watermarks degrade the performance of \vlm in document understanding?} To address this, we propose a novel evaluation framework to investigate the effect of visible watermarks on \vlm performance.
We takes into account various factors, including different types of document data, the positions of watermarks within documents and variations in watermark content.
Our experimental results reveal that \vlm  performance can be significantly compromised by watermarks, with performance drop rates reaching up to 36\%. We discover that \emph{scattered} watermarks cause stronger interference than centralized ones, and that \emph{semantic contents} in watermarks creates greater disruption than simple visual occlusion. Through attention mechanism analysis and embedding similarity examination, we find that the performance drops are mainly attributed to that watermarks 1) force widespread attention redistribution, and 2) alter semantic representation in the embedding space. Our research not only highlights significant challenges in deploying \vlm for document understanding, but also provides insights towards developing robust inference mechanisms on watermarked documents.



\end{abstract}

\section{Introduction}

The rapid advancement of Visual Language Models (\vlm) has transformed \du capabilities, enabling sophisticated processing of complex multimodal content without relying on traditional Optical Character Recognition (\ocr) technologies. 
Unlike conventional document processing systems that convert text content into machine-readable characters \citep{Srihari1986DocumentIU, hwang2021spatialdependencyparsingsemistructured, hong2022brospretrainedlanguagemodel}, modern \vlm directly process original document images, seamlessly integrating visual features with textual information. This OCR-free approach eliminates potential recognition errors in documents with complex layouts, non-standard fonts, and visual noise \citep{kim2022ocrfreedocumentunderstandingtransformer, liu2024textmonkeyocrfreelargemultimodal, hu2024mplugdocowl2}, while preserving critical layout and semantic structures that enhance performance in tasks such as question answering, information retrieval, and cross-page analysis.

However, this advancement introduces new robustness challenges that remain largely unexplored. While considerable research has examined the vulnerability of \vlm to adversarial inputs and visual noise in general contexts, the specific impact of common document modifications, particularly watermarks on \vlm performance, represents a critical gap in our understanding. Unlike Large Language Models (\llm), where robustness mechanisms against harmful queries are relatively well-established \citep{mei2024not}, \vlm' multimodal nature creates unique vulnerability surfaces where image features can potentially bypass or undermine existing safeguards.

\begin{figure*}
    \centering
    \includegraphics[width=1 \textwidth]{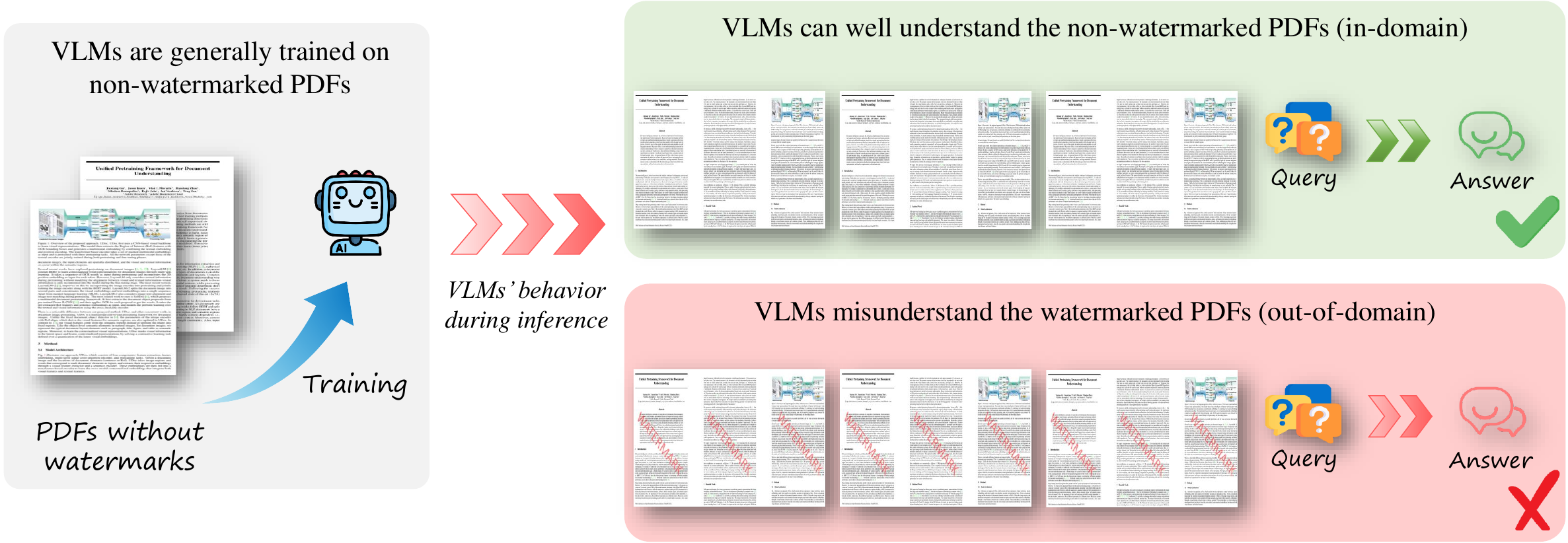}
    \caption{After being fine-tuned on unwatermarked document data, the LVLM provides correct responses to question with unwatermaked document but exhibits obvious errors when responding to watermarked one.}
    \label{fig:ewatermarkexample}
\end{figure*}

Watermarking represents one of the most prevalent forms of visual modification in professional documents, serving as a standard method for copyright protection and document authentication \citep{779263}. 
As illustrated in \Cref{fig:ewatermarkexample}, even a simple watermark can dramatically alter an \vlm' document interpretation, causing the model to produce incorrect responses despite no changes to the underlying informational content. This observation raises fundamental questions about \vlm reliability in real-world document processing scenarios, where watermarked materials are the norm rather than the exception.

To address this critical knowledge gap, we introduce a comprehensive evaluation framework specifically designed to assess \vlm robustness against watermarks in document understanding tasks. 
Our approach considers the unique characteristics of document-based question answering across diverse content types, including text-heavy documents, charts, and tables \citep{suri2024visdommultidocumentqavisually}. By systematically embedding visible watermarks with varying properties into clean documents, we generate a watermarked dataset and the original clean dataset.
Each dataset entry comprises a document image
\footnote{Documents exist in various formats, such as DOC, PDF, and PPT, however, images provide a standardized representation. 
} 
accompanied by a corresponding question.  
Our evaluation is conducted on four state-of-the-art \vlm.
The model performance is assessed across the two datasets, providing insights into the model's sensitivity to visible watermarks.
The primary objectives of our study are as follows:  
\circled{1} To analyze the impact of visible watermarks on the performance of \vlm for \du.  
\circled{2} To identify the key characteristics of visible watermarks that most significantly influence model robustness.


With the proposed framework, we perform numerous controlled experiments to find that
increasing watermark size does not necessarily amplify the interference effect, contradicting our initial hypothesis. Further analysis reveals that: (1) For watermarks with the same total area, distributing them across multiple locations results in significantly stronger interference than placing them in a single fixed position; (2) By examining watermarked documents that successfully cause perturbation, we observe that effective watermarks are often located near the region containing the standard answer, suggesting that obstruction of key content leads to interference; (3) To validate this `occlusion' reason, we replace text watermark `MARK' with rectangular mask to improve the probability of covering key information and repeat the experiments. However, the interference effect is notably weaker compared to text-based watermarks,
highlighting the importance of \textit{semantic information} in watermarks rather than `occlusion' and provide insights for designing more robust watermarking strategies.

We analyze the impact of watermark position and content from two perspectives: attention mechanisms and semantic similarity in embedding space. Our findings show that:  
(1) The most disruptive watermark position causes the widest redistribution of attention weights.  
(2) The most disruptive watermark content results in the lowest semantic similarity between the embeddings of watermarked and original documents.

This paper makes several contributions to the literature:
First, we systematically study and evaluate the vulnerabilities of \vlm for \du for the first time.
Second, we propose a new framework to evaluate the impact of visual watermarks on \vlm performance. 
Third, we analyze how watermark position and content affect \vlm performance, and use visualization to reveal the underlying causes.

\section{Related Works}
\subsection{Visual Language Models for Document Understanding}
The field of \vdu focuses on understanding structure in visual documents and extracting key information to convert structured or semi-structured forms into machine-readable formats. A visual document is a document that contains a variety of visual and text elements. These documents integrate a variety of visual elements, including paragraphs, charts, tables, etc. Text elements (such as paragraphs and lists) and visual elements (such as charts and tables) together constitute the semantics of the documents. \qa is one of the key tasks in visual document understanding, which involves answering natural language questions based on the context information of visual documents\citep{ding2024deeplearningbasedvisually}.

With the rapid development of visual document understanding requirements, \vlm technology for visual document understanding is introduced: 
Qwen-VL \citep{bai2023qwenvlversatilevisionlanguagemodel} retains the Q-Former architecture while replacing the original language model with a larger one. 
In order to extend the capability of \vlm to \vdu,  mPLUG-DocOwl \citep{ye2023mplugdocowl} proposes a modular model based on mPLUG-Owl \citep{ye2023mplugowlmodularizationempowerslarge} 
for \du without \ocr. 
These comprehensive approaches highlight the potential of \vlm in handling complex visual language tasks \citep{Li_2024_CVPR}.

\subsection{Robustness of Visual Language Models} \label{sec:robustness}
With the widespread application and impressive performance of \vlm in multimodal \qa and reasoning tasks, their robustness has garnered increasing attention in recent years.
\citet{zhao2023evaluatingadversarialrobustnesslarge} evaluate the adversarial robustness of \vlm and find that multimodal generation increases security risks and that an attacker can deceive the model by manipulating visual input.
\citet{qiu2024benchmarkingrobustnessmultimodalimagetext} observe that the multimodal model is not robust to image and text perturbation, especially to image perturbation. In the perturbation method tested, character-level perturbation causes the most serious distribution deviation of text, and zoom blur is the most serious deviation of image data.
\citet{lee2024vhelmholisticevaluationvision} introduce VHELM to evaluate the robustness of \vlm through a variety of test scenarios, such as introducing typos and testing robustness to sketches and out-of-distribution images.
\citet{agarwal2025mvtamperbenchevaluatingrobustnessvisionlanguage} introduce MVTamperBench, designed to evaluate the robustness of \vlm against video tampering effects such as rotation, occlusion, substitution, and repetition. 


Significant progress has been achieved in robustness research of \vlm. In this work, we focus on how visual watermarks influence \vlm performance when applied to document understanding, identifying potential risks in visual question answering scenarios.
\section{Methodology}\label{sec:exp}
In this section, we describe our methodology for evaluating the impact of watermarks on \vlm in document understanding tasks. We first formalize the document understanding task and define performance metrics to quantify model accuracy. 
Following this, we detail our evaluation datasets and experimental setup, including the selection of models and watermark parameters.

\subsection{Document Understanding Task and \vlm Performance Metric}
We use visual question answering (\vqa) as our primary task.
Because \vqa is the most essential and widely used task in document understanding scenarios.
\citep{li2023evaluatinginstructionfollowingrobustnesslarge}.  
The \vqa task is defined as follows: 

\noindent
Given a user question $q$ and a document image $i$ as visual context, the system is tasked with generating an answer. Each question is accompanied by a corresponding ground truth $gt$, which serves as the reference for evaluation. Our evaluation dataset $\mathcal{D}_{\text{eval}}$ consists of representative \vqa data, with each sample structured as $(q,i,gt)$.
For a given \textsc{VLM} model $f$, which takes the question-image pair $(q, i)$ as input and generates an answer, the \textit{answer accuracy} over $\mathcal{D}_{\text{eval}}$ is defined as:  
\begin{equation*}
\text{Acc}(f, \mathcal{D}_{\text{eval}}) \overset{def}{=} \frac{1}{\left | \mathcal{D}_{\text{eval}}\right |}\sum_{(q,i,gt)\in \mathcal{D}_{\text{eval}}}\mathbbm{1}(f(q,i), gt),
\end{equation*}
\label{v-meaning}

where $\mathbbm{1}(\cdot)$ is the indicator function, which takes the value of 1 if the model's generated answer $f(q,i)$ matches $gt$, and 0 otherwise. 


To evaluate the impact of the watermark on the model’s performance, we inject a watermark $w$ into each input image $i$ in $\mathcal{D}_{\text{eval}}$, resulting in a perturbed evaluation dataset, denoted as $\mathcal{D}_{\text{eval}}'$, which consists of samples in the form of $(q,i+w,gt)$.

To quantify this impact, we adopt the \textbf{Performance Drop Rate (PDR)} metric \citep{zhu2024promptrobustevaluatingrobustnesslarge}, 
which measures the percentage decrease in \textit{answer\ accuracy} with respect to the user question $q$. 
The PDR of \textsc{VLM} $f$ under the influence of the watermark is defined as:

\begin{equation*}
\text{PDR}(f, \mathcal{D}_{\text{eval}}, \mathcal{D}_{\text{eval}}') \overset{def}{=} 
\frac{
\text{Acc}(f, \mathcal{D}_{\text{eval}})
- 
\text{Acc}(f, \mathcal{D}_{\text{eval}}')
}
{
\text{Acc}(f, \mathcal{D}_{\text{eval}})
}.
\end{equation*}

\noindent
The PDR metric directly captures the degradation in model performance when processing watermarked documents compared to original documents, providing a clear measure of how significantly watermarks interfere with the model's document understanding capabilities.
A higher PDR value indicates a stronger disruptive effect of watermarks on the model.



\begin{figure*}
    \centering
    \includegraphics[width=1 \textwidth, height=0.35 \textheight]{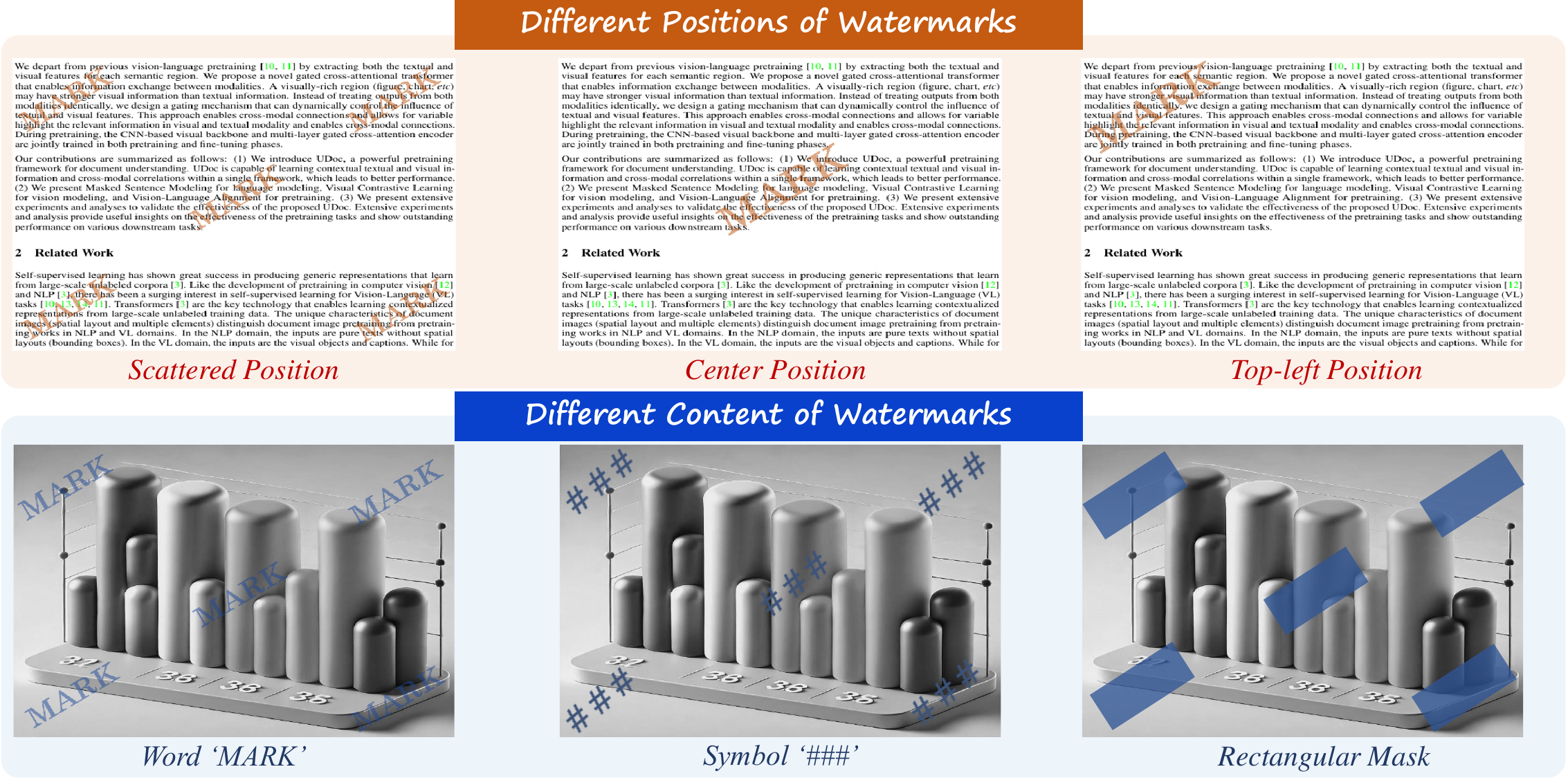}
    \caption{\textbf{Position and content exploration.} (a) For position exploration, the watermark is placed \sca in five fixed positions, in \cen and in \topleft corner of document image; (b) For content exploration, we set the watermark content to text \markk, \symboll, and fill rectangular box where the watermark is located with a rectangular \mask.}
    \label{fig:exploration}
\end{figure*}

\subsection{Experimental Setting}

\paragraph{Evaluation Dataset} \label{para:dataset}
We construct our datasets by randomly sampling 100 images for each of the three document types — text, chart, and table — from the public \docowl\ dataset \citep{hu2024mplugdocowl2} and the LongDocURL dataset available on Hugging Face \citep{dengchao2024longdocurl}.
These images are categorized into four classes: TextS, ChartS, ChartM, and TableS, corresponding to text document images, chart document images, and table document images, along with their respective question types (single choice or multiple response).

For each image, we design a set of questions based on its content, providing four answer choices for the model to select from. The number of correct answers is determined by the category of the image: images in the ChartM class have questions with multiple correct answers, whereas images in the TextS, ChartS, and TableS classes have questions requiring only a single correct choice.
The resulting four datasets are as follows:


\begin{itemize}  
    \item \textbf{TextS Set}: Each instance in this dataset consists of a text document image paired with a corresponding single-choice question. Each question presents four answer options, and the groundtruth is a single correct option among them.  
    \item \textbf{ChartS Set}: Each instance in this dataset consists of a chart document image paired with a corresponding single-choice question. Each question includes four answer choices, with only one correct answer designated as the groundtruth.  
    \item \textbf{ChartM Set}: Each instance in this set consists of a chart document image paired with a multiple-response question. Each question provides four answer choices, and the groundtruth comprises more than one correct options among them.  
    \item \textbf{TableS Set}: Each instance in this set consists of a table document image paired with a single-choice question. Each question contains four possible answers, with only one correct option specified as the groundtruth.  
\end{itemize}

For text-based documents, 50\% are from real estate and finance, and 50\% from law-related studies. Chart and table documents are extracted from academic papers in computer science, physics, and related fields.
This diverse collection of document types and question formats enables us to comprehensively evaluate how watermarks affect \vlm across different document understanding scenarios, from simple text extraction to complex chart interpretation and tabular data analysis.

\paragraph{Evaluation Models.} 
We select 4 representative sate-of-the-art \vlm for our evaluation: 
\docowl \citep{hu2024mplugdocowl2} 
is designed specifically for \du tasks, employing a specialized architecture for processing document layouts and mixed content types.\\
\internvl 
\citep{chen2025expandingperformanceboundariesopensource}
excels in real-world understanding benchmarks and is capable of handling high-resolution images and complex document layouts.\\
\llava 
\citep{liu2024improvedbaselinesvisualinstruction}
features an attention mechanism that dynamically focuses on key parts of text or images while processing document content.\\
\qwen 
\citep{bai2025qwen25vltechnicalreport}
demonstrates strong performance in document understanding tasks, particularly in structured data extraction and long document understanding.\\
This selection covers a range of model architectures and specializations, allowing us to identify both general trends and model-specific behaviors in response to watermarked documents.

\paragraph{Experimental Parameters.}
We primarily investigate the impact of watermark position and content on the interference effect. All watermarks are programmatically added to the document images using the Python Pillow library. To ensure experimental consistency and isolate the effects of other watermark properties, we keep several key parameters fixed:  (1) We set watermark color as black and set the watermark transparency to 0.5 to ensure that the existence of the watermarks does not affect the visibility of the document to humans, but also has a certain impact on the performance of the model, which is more in line with the realistic scene of document watermarks; (2) We fix the watermark angle at 0 degrees to eliminate orientation as a variable;  (3) Additionally, in each experimental condition, the proportion of watermarks in the total area of the document image is distributed between 10\% and 80\%.
Experiments on watermark angle and opacity are included in the appendix.

\paragraph{Considered Watermark Properties.}
Our experiments focus on two primary dimensions of watermark properties, examining how each affects \vlm performance (see \Cref{fig:exploration}):\\
(1) \textbf{Watermark Position}: We systematically investigate how watermark placement influences \vlm performance by positioning watermarks in three distinct configurations. The \textsc{Center} placement positions the watermark at the center of the document image. The \textsc{Top-left} configuration places the watermark in the top-left corner of the document. The \textsc{Scattered} approach distributes watermarks across five fixed positions throughout the document.\\
(2) \textbf{Watermark Content}: We evaluate how different types of watermark content affect \vlm performance using three variants. The \textsc{MARK} content consists of a text-based watermark containing the word `MARK'. The \textsc{`\#\#\#'} uses a symbol-based watermark with the pattern \textsc{`\#\#\#'}. The \textsc{Mask} variant employs a rectangular mask placed in the same location as the text-based watermark.
\\
For both dimensions, we carefully control all variables except the one being explored. This ensures that any observed differences in performance can be directly attributed to the specific watermark property under investigation. The total watermark area, color, transparency, and other factors remain consistent across experimental conditions within each dimension.

\section{Impact of Document Watermarks on \vlm Performance}

\begin{table}[t]
\begin{center}
\setlength{\tabcolsep}{1pt}
  \tiny
      \caption{
\textbf{PDR($\%$) for different \textsc{watermark positions}}. Note: The \colorbox{myred}{pink numbers} represent the highest PDR value across different watermark positions.}
\label{table:main-position}
\begin{tabular}{l ccc| ccc| ccc| ccc}
\toprule \multirow{2}*{Model}
         & \multicolumn{3}{c}{\docs}  
         & \multicolumn{3}{c}{\charts} 
         & \multicolumn{3}{c}{\chartm} 
         & \multicolumn{3}{c}{\tables}
         \\

\cmidrule(lr){2-4}
\cmidrule(lr){5-7}
\cmidrule(lr){8-10}
\cmidrule(lr){11-13}

         & \cen & \sca & \topleft
         & \cen & \sca & \topleft
         & \cen & \sca & \topleft
         & \cen & \sca & \topleft \\
\midrule

\docowl & 12 & \cellcolor[HTML]{ffe1ee}14 & 1 & 1 & \cellcolor[HTML]{ffe1ee}11 & 3 & 15 & \cellcolor[HTML]{ffe1ee}24 & 9 & 24 & \cellcolor[HTML]{ffe1ee}28 & 1 \\
\midrule
\internvl-8b & 21 & \cellcolor[HTML]{ffe1ee}28 & 16 & \cellcolor[HTML]{ffe1ee}2 & \cellcolor[HTML]{ffe1ee}2 & 1 & 25 & \cellcolor[HTML]{ffe1ee}36 & 28 & 10 & 11 & \cellcolor[HTML]{ffe1ee}14 \\
\midrule
\llava-7b & \cellcolor[HTML]{ffe1ee}19 & 9 & 11 & 9 & \cellcolor[HTML]{ffe1ee}23 & 20 & 16 & \cellcolor[HTML]{ffe1ee}22 & 21 & \cellcolor[HTML]{ffe1ee}21 & 17 & 10 \\
\midrule
\qwen-7b & 0 & 0 & 0 & 4 & \cellcolor[HTML]{ffe1ee}6 & 5 & 1 & \cellcolor[HTML]{ffe1ee}7 & 2 & 0 & 0 & 0 \\
\midrule
\midrule
AVG & \cellcolor[HTML]{ffe1ee}13 & 12 & 7 & 4 & \cellcolor[HTML]{ffe1ee}10 & 7 & 14 & \cellcolor[HTML]{ffe1ee}22 & 15 & 13 & \cellcolor[HTML]{ffe1ee}14 & 6 \\
\bottomrule
  \end{tabular}
  \end{center}

\end{table}

\Cref{table:main-position} and \Cref{table:main-content} show \vlm $\text{PDR}(f, \mathcal{D}_{\text{eval}}, \mathcal{D}_{\text{eval}}')$ on datasets with watermarks of different positions and contents.
In general, the highest PDR can reach 36\%, indicating that \vlm are susceptible to the interference of document watermarks when processing document understanding tasks.

What's more, we investigate the impact of watermark color and area ratio on model performance.
Results (see appendix) show that:
(1) Color change doesn't significantly affect \vlm performance.
This observation can be attributed to how modern \vlm process textual and visual information.
These models primarily focus on high-level semantic understanding rather than low-level pixel variations. 
(2) A larger watermark area ratio does not necessarily increase interference. 
This is because when the watermark area is too large, the semantic information carried by the watermark becomes diluted, making it ineffective in interfering with document information.

\subsection{Impact of \textsc{Watermark Position}}

\subsubsection{PDR across Watermark Positions}

We can observe from \Cref{table:main-position} that the interference effect of watermarks is significantly influenced by their position.  \\
(1) \textbf{Position Effect.} 
 Watermarks of \sca configuration consistently result in higher PDR values across most testing cases compared to those of the \cen and \topleft configurations. In 11/16 cases, the \sca position yields the highest PDR among the three locations, demonstrating its dominant impact. For \docowl on the \chartm dataset, \sca watermarks achieve 24\% PDR compared to 15\% for \cen and 9\% for \topleft placements. This pattern is particularly evident in the \charts and \chartm datasets, where \sca watermarks consistently lead to the highest PDR values across all tested models.
\\
(2) \textbf{Dataset Vulnerability. } 
The overall PDR levels vary significantly across datasets. The \chartm dataset exhibits the highest PDR values, where it reaches up to 36\% (\internvl of \sca configuration).
This suggests that the \chartm dataset, which by nature allows for multiple correct answers, already presents a challenge to model performance.
The presence of watermarks exacerbates this issue, further reducing the model's prediction accuracy.
\\
(3) \textbf{Model Robustness.} 
Different \vlm exhibit varying sensitivity to watermarks. \qwen consistently demonstrates low PDR values, with PDR values even dropping to zero on the \docs and \tables datasets, indicating its strong robustness against watermarks. In contrast, \internvl and \docowl show higher vulnerability to watermark interference, especially on chart-based documents.\\


\subsubsection{Attention Mechanism Analysis}

\begin{figure}
    \centering
    \includegraphics[width=1\linewidth]{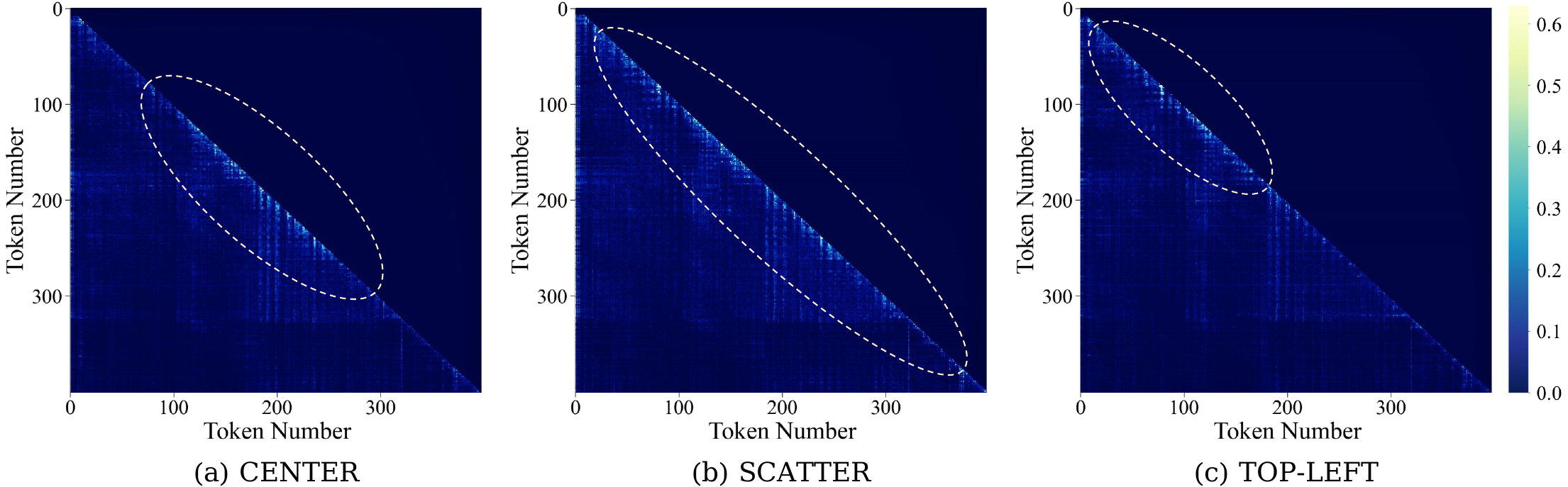}
    \caption{\textbf{Heatmap of attention wight variation} between watermarked multimodal input sequences and original one. 
    \colorbox{myblue}{\texttt{Brighter points}} indicate the model's attention weights for the corresponding tokens have changed, with higher brightness indicating greater changes. 
    The \colorbox{myyellow}{\texttt{yellow dashed line}} indicates the region where the attention weight changes most significantly.
    }
    \label{fig:attention-weight}
\end{figure}

To further explore the underlying reasons why watermarks with different positions produce different interference effects, 
we calculate the variation in \emph{attention weights} generated by \docowl model between the watermarked multimodal input sequence and the non-watermarked input sequence, and use it to produce the attention weight variation heatmap (see \Cref{fig:attention-weight}).
The same analysis is also conducted for \qwen and \llava, with results in the appendix.
The visualization reveals distinct patterns:
the model adjusts \emph{attention weights} across the entire image under the \sca setting (see \Cref{fig:attention-weight}(b)), while attention changes are concentrated in the middle (see \Cref{fig:attention-weight}(a)) and beginning portions (see \Cref{fig:attention-weight}(c)) of the sequence under the \cen and \topleft settings, respectively.


These attention variations can be attributed to the VLM's processing architecture. Before performing multimodal fusion, the model divides the image into different patches. When watermarks are embedded in different positions, they affect different patches in the input sequence, altering the embedding vectors at these specific locations. These changes subsequently induce corresponding modifications in the \emph{attention weights} at these positions.

The key insight from these results is that \sca watermarks disrupt the model's attention across multiple regions, forcing widespread \emph{attention redistribution} that compromises the model's ability to maintain coherent document understanding. In contrast, when watermarks are concentrated in a single location (\cen or \topleft), the model primarily adjusts its attention in that affected region while preserving its processing of information elsewhere. This localized adaptation allows the model's \emph{semantic understanding} to remain largely intact, resulting in lower performance degradation compared to the \sca configuration.


\subsection{Impact of \textsc{Watermark Content}}

After examining position effects, we now investigate how different watermark content types affect \vlm performance. While position determines where interference occurs, content characteristics may influence how the model processes and interprets the watermarked information. We test three distinct watermark content: text-based ('MARK'), symbol-based ('\#\#\#'), and visual mask (MASK) to understand their varying impacts on document understanding capabilities.

\begin{table}[t]
\begin{center}
\setlength{\tabcolsep}{1pt}
  \tiny
\caption{
\textbf{PDR($\%$) for different \textsc{watermark content}}. Note: The \colorbox{mypurple}{purple numbers} represent the highest PDR value across different watermark content.}
  \label{table:main-content}
\begin{tabular}{l ccc| ccc| ccc| ccc}
\toprule \multirow{2}*{Model}
         & \multicolumn{3}{c}{\docs}  
         & \multicolumn{3}{c}{\charts} 
         & \multicolumn{3}{c}{\chartm} 
         & \multicolumn{3}{c}{\tables}\\

\cmidrule(lr){2-4}
\cmidrule(lr){5-7}
\cmidrule(lr){8-10}
\cmidrule(lr){11-13}

         & \markk & \symboll & \mask
         & \markk & \symboll & \mask
         & \markk & \symboll & \mask
         & \markk & \symboll & \mask \\
\midrule

\docowl & 6 & 8 & \cellcolor[HTML]{e9e1ff}16 & \cellcolor[HTML]{e9e1ff}11 & 6 & 1 & \cellcolor[HTML]{e9e1ff}28 & 10 & 9 & \cellcolor[HTML]{e9e1ff}33 & 32 & 7 \\
\midrule
\internvl-8b & 22 & \cellcolor[HTML]{e9e1ff}25 & 17 & \cellcolor[HTML]{e9e1ff}2 & 1 & 1 & \cellcolor[HTML]{e9e1ff}35 & 26 & 28 & \cellcolor[HTML]{e9e1ff}14 & 9 & 11 \\
\midrule
\llava-7b & 16 & \cellcolor[HTML]{e9e1ff}21 & 3 & \cellcolor[HTML]{e9e1ff}22 & 17 & 13 & 12 & \cellcolor[HTML]{e9e1ff}23 & \cellcolor[HTML]{e9e1ff}23 & \cellcolor[HTML]{e9e1ff}18 & 16 & 14 \\
\midrule
\qwen-7b & 0 & 0 & 0 & 6 & \cellcolor[HTML]{e9e1ff}8 & 0 & 2 & \cellcolor[HTML]{e9e1ff}5 & 2 & 0 & 0 & 0 \\
\midrule
\midrule
AVG & 11 & \cellcolor[HTML]{e9e1ff}13 & 9 & \cellcolor[HTML]{e9e1ff}10 & 8 & 4 & \cellcolor[HTML]{e9e1ff}19 & 16 & 16 & \cellcolor[HTML]{e9e1ff}16 & 14 & 8 \\

\bottomrule
  \end{tabular}
  \end{center}
\end{table}

\subsubsection{PDR across Watermark Content}
As shown in \Cref{table:main-content}, watermark content significantly influences interference effects across models and datasets.

(1) \textbf{Content Effect.} 
When the watermark content is either \markk or \symboll, the PDR values of \vlm are generally higher compared to the \mask watermarks. 
From a statistical perspective, the average PDR across all models and datasets for the \markk setting reaches 14\%, for the \symboll setting 13\%, and for the \mask setting only 9\%. 
In most cases, the PDR under the \markk setting are highly similar to that under the \symboll setting, while the \mask setting exhibits a noticeably lower and more varied interference effect. 
These results highlight that different watermark content types impact document understanding tasks to varying degrees, with explicit textual and symbolic watermarks being the most disruptive across multiple models and datasets.

(2) \textbf{Model Robustness.} 
Different \vlm show varying sensitivity patterns to watermark content. \internvl and \llava are more susceptible to interference from \symboll watermarks on the \docs dataset, whereas they are more affected by \markk watermarks on other datasets. 
Similarly, the \qwen model shows sensitivity to watermarks only on the \charts and \chartm datasets, with the highest PDR in the \symboll configuration. 
Additionally, the \docowl model exhibits significant sensitivity to \mask watermarks on the \docs dataset, with its PDR values for \markk and \symboll watermarks being substantially lower than those for \mask.


\subsubsection{Semantic Interference Analysis}
The experimental results indicate that the impact of watermarks on model performance is not solely due to their obstruction of document information. If the primary effect were merely occlusion, the \mask watermarks should outperform the \markk and \symboll watermarks. However, our findings do not fully support this assumption. Therefore, we further analyze the influence of watermark content from the perspective of \emph{semantic alterations} in the \emph{cross-modal embedding} of input sequence. 

\begin{figure*}
    \centering
    \begin{minipage}{0.45\textwidth} 
        \centering
        \includegraphics[width=1\textwidth]{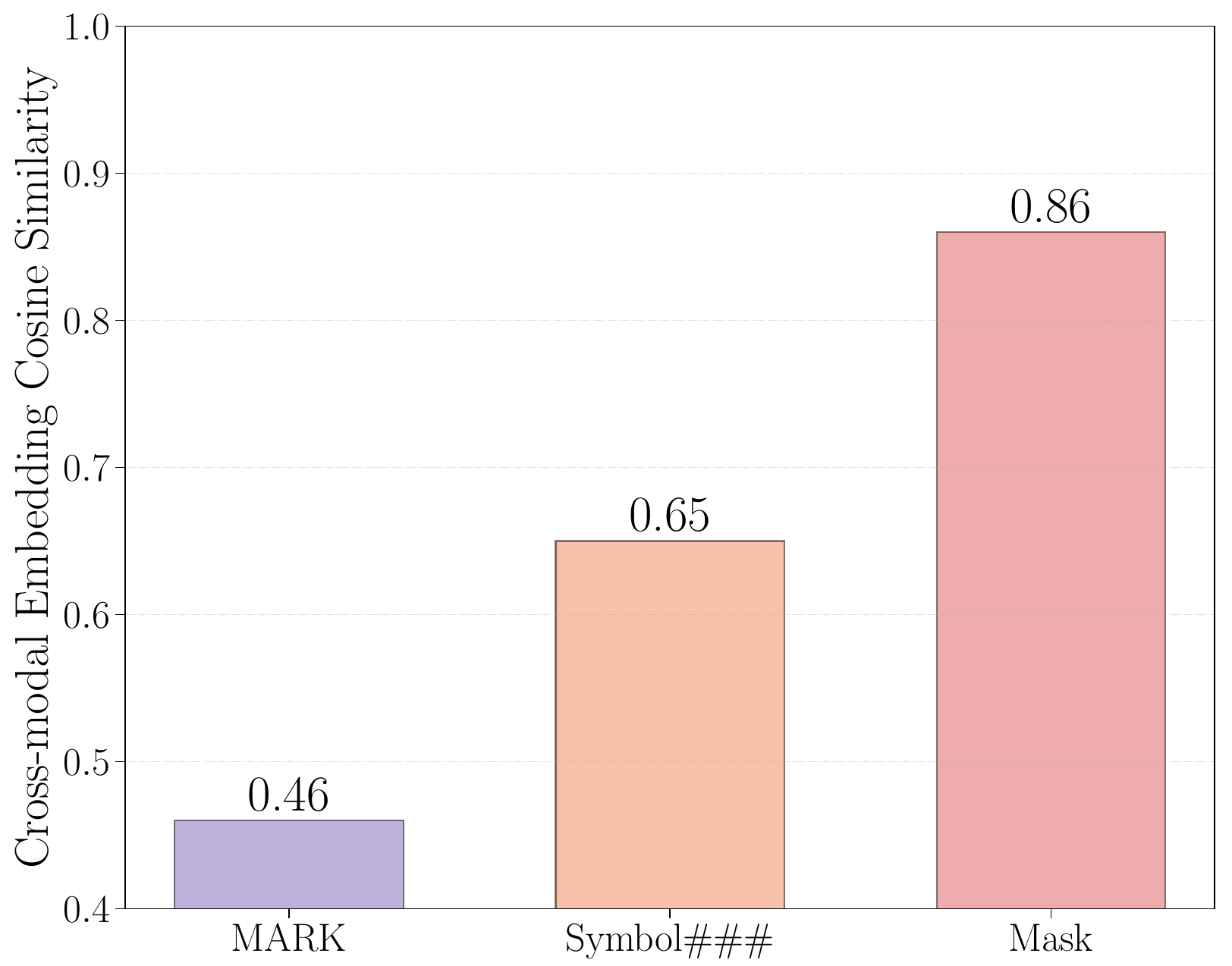}
        \caption{\textbf{Cross-modal embedding cosine similarity} of watermarked input sequences with the original one.
        }
        \label{fig:cosinesim}
    \end{minipage}
    \hfill
    \begin{minipage}{0.52\textwidth} 
        \centering
        \includegraphics[height=0.22\textheight]{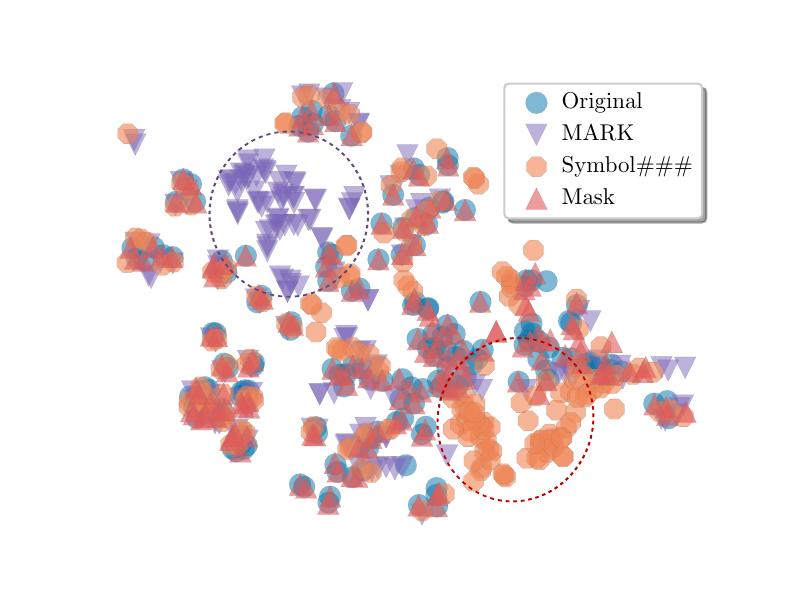}
        \caption{Dimentional \textbf{t-SNE visualization} of input cross-modal embeddings with watermarks in different content.}
        \label{fig:t-sne}
    \end{minipage}
\end{figure*}

\textbf{Cross-modal Embedding Cosine Similarity.} 
\emph{Cosine similarity} evaluates the similarity between two embedding vectors in the semantic space by measuring the directional similarity between them \citep{zhu2025mitigating, 10.1145/361219.361220}.
The higher the cosine similarity score is, the more similar the two embedding vectors are.
We compute the cosine similarity between the \emph{cross-modal embedding} vectors obtained from the watermarked and non-watermarked input sequence using the \docowl model (see \Cref{fig:cosinesim}).
The results reveal that the \markk watermark leads to the lowest cosine similarity with the original input (0.46), whereas the \mask watermark results in the highest similarity (0.86). This observation further confirms that although the \mask watermark maximizes the occlusion of key information, it induces minimal changes in the cross-modal  embeddings' semantic information. Consequently, this can explain why the average PDR value under the \mask setting is the lowest.

\textbf{Dimentional t-SNE Visualization for Cross-modal Embedding.} 
t-SNE \citep{hinton2002stochastic} can be used to compare the distribution characteristics of \emph{cross-modal embedding}s before and after watermarking. If the addition of  watermarks causes a significant change in the embedding distribution, it indicates that watermarks have affected the model's semantic understanding. 
As shown in \Cref{fig:t-sne} , the projections of \emph{cross-modal embedding}s (generated from \docowl) with \markk and \symboll watermarks in the two-dimensional space exhibit a noticeable deviation from the original input data points, whereas embedding with \mask watermarks remain highly consistent with the original one, showing almost no significant change. 
This phenomenon suggests that \markk and \symboll watermarks introduce additional explicit information, altering semantic structure of the input content, which in turn leads to a substantial shift in the \emph{cross-modal embedding}. 
In contrast, although \mask watermarks visually obscure part of the input content, they don't add new semantic information. 
As a result, the model can maintain a relatively stable embedding distribution during cross-modal alignment. 
This also explains why the PDR values for \markk and \symboll watermark settings are more similar, while the PDR for the \mask watermark differs certainly from the other two.

\section{Conclusion}


In this paper, we establish a framework based on \vqa datasets to systematically evaluate the robustness of \vlm in \du task for the first time. 
Through comprehensive experiments on four state-of-the-art \vlm, we observe that watermarks in documents can introduce varying degrees of interference in model performance. 
Our study provides new insights into robustness evaluation for \vlm in \du tasks, highlighting the need for further research in assessing and enhancing the robustness of \vlm in multimodal document processing. 

\newpage
\bibliography{colm2025_conference}
\bibliographystyle{colm2025_conference}

\appendix
\newpage
\section{Appendix}
\label{pg:appendix}


\subsection{Impact of Watermark Color}
We compare the PDR across datasets with different watermark colors on \docowl, all positioned in a \sca position. The results, shown in \Cref{table:color}, indicate that average PDR scores across different colors are similar. 
This observation can be attributed to how modern \vlm process textual and visual information. These models primarily focus on high-level semantic understanding rather than low-level pixel variations. As a result, changes in watermark color do not significantly affect the model’s interpretation of the document, leading to minimal differences in interference effects.

\begin{table}[h]
    \centering
    \setlength{\tabcolsep}{1pt}
    \tiny
    \caption{\textbf{Average PDR($\%$) for different \textsc{watermark colors} on \docowl.}}
    \resizebox{\linewidth}{!}{
        \begin{tabular}{l ccc| ccc| ccc| ccc c}
        \toprule 
        \multirow{2}*{Color}
                 & \multicolumn{3}{c}{\docs}  
                 & \multicolumn{3}{c}{\charts} 
                 & \multicolumn{3}{c}{\chartm} 
                 & \multicolumn{3}{c}{\tables}
                 & \multirow{2}*{AVG}\\

        \cmidrule(lr){2-4}
        \cmidrule(lr){5-7}
        \cmidrule(lr){8-10}
        \cmidrule(lr){11-13}

                 & \markk & \symboll & \mask
                 & \markk & \symboll & \mask
                 & \markk & \symboll & \mask
                 & \markk & \symboll & \mask \\
        \midrule

        \textsc{Black} & 11 & 7 & 21 & 20 & 10 & 2 & 55 & 9 & 5 & 39 & 38 & 76 & 19 \\
        \midrule
        \textsc{Red} & 0 & 0 & 32 & 18 & 10 & 1 & 63 & 33 & 4 & 31 & 35 & 4 & 20 \\
        \midrule
        \textsc{Green} & 0 & 1 & 30 & 7 & 11 & 0 & 64 & 20 & 2 & 30 & 40 & 3  & 22\\

        \bottomrule
        \end{tabular}
    }
    \label{table:color}
\end{table}

\noindent
\subsection{Impact of Watermark Area Ratio}
\begin{figure}[h]
    \centering
    \includegraphics[width=0.8\linewidth]{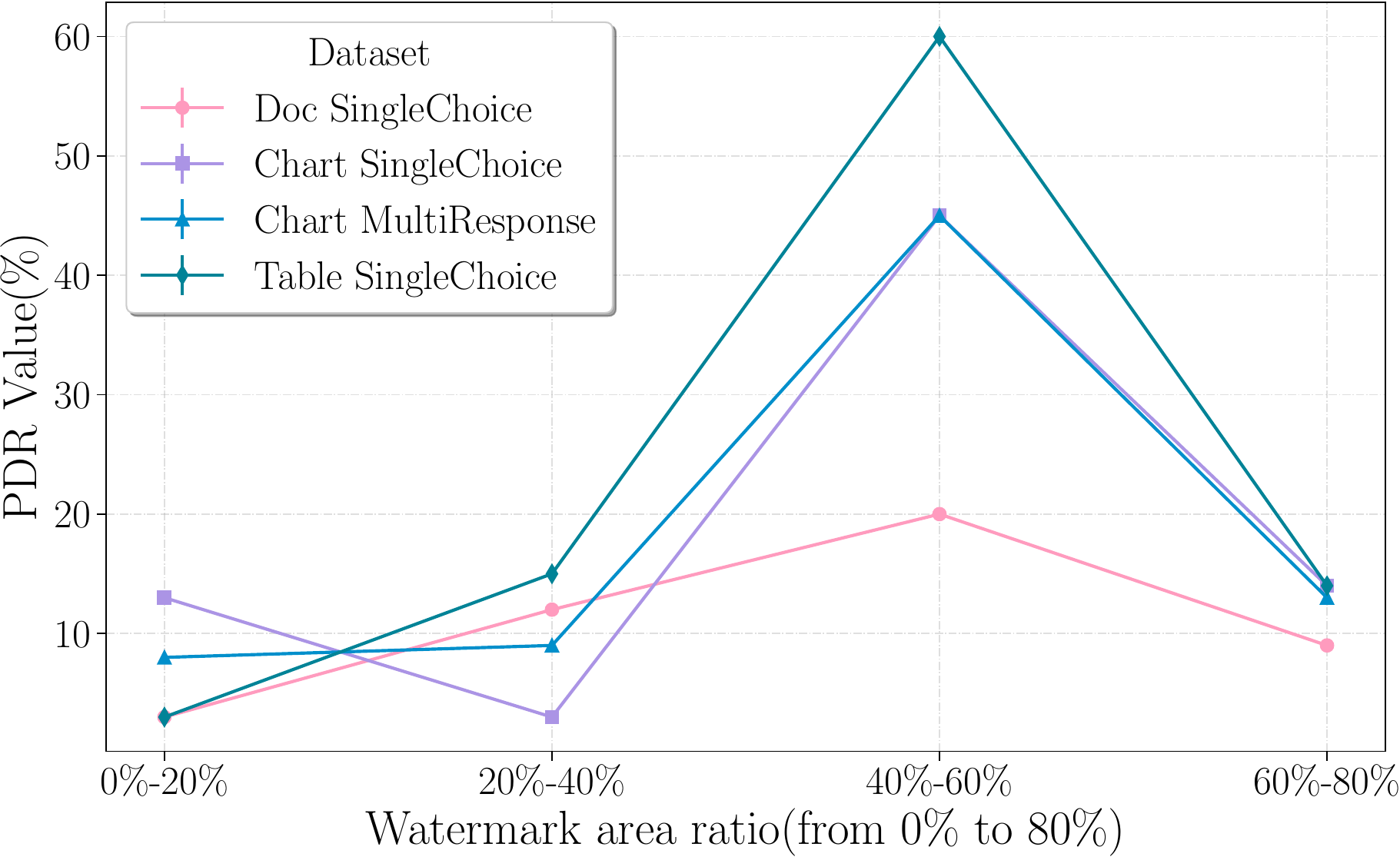}
    \caption{\textbf{Average PDR(\%) of different watermark area ratios} on the full dataset.}
    \label{fig:arearatio}
\end{figure}

As shown in \Cref{fig:arearatio}, increasing the area 
ratio of the watermark does not necessarily lead to a stronger interference effect. This suggests that once the watermark area exceeds a certain threshold, it is no longer the decisive factor influencing model performance. This can be attributed to that when the watermark area is too large, the semantic information carried by the watermark becomes diluted, making it ineffective in interfering with the original content of the document.

\subsection{Impact of Watermark Opacity}

As shown in \Cref{tab:opacity}, increasing the opacity of the watermark leads to a significant rise in performance degradation, with the average PDR increasing from 8.5\% at low opacity ($\alpha=0.2$) to 21.3\% at high opacity ($\alpha=0.8$). This trend suggests that higher visual prominence caused by denser watermark opacity results in stronger disruption of model attention. The underlying reason may be that more opaque watermarks dominate the visual space, increasing misalignment in the model’s attention mechanism and impairing its ability to focus on relevant document content.

\begin{table}[h]
\centering
\caption{Average PDR(\%) for different watermark opacity levels on mPLUG-DocOwl2.}
\begin{tabular}{lc}
\toprule
\textbf{Opacity ($\alpha$)} & \textbf{Avg. PDR (\%)} \\
\midrule
0.2 (light)   & 8.5  \\
0.5 (medium)  & 15.7 \\
0.8 (dense)   & 21.3 \\
\bottomrule
\end{tabular}
\label{tab:opacity}
\end{table}

\subsection{Impact of Watermark Rotation}
As shown in \Cref{tab:rotation}, variations in watermark rotation angle have minimal effect on model performance. The average PDR remains relatively stable across different angles, with 14.2\% at $0^\circ$, 14.5\% at $45^\circ$, and 14.1\% at $90^\circ$. This consistency indicates that modern VLMs exhibit strong robustness to moderate geometric transformations in watermark orientation. A possible explanation is that the spatial encoding mechanisms of current vision backbones effectively normalize such rotation variations, making them less disruptive to cross-modal alignment.

\begin{table}[h]
\centering
\caption{Average PDR(\%) for different watermark rotation angles on mPLUG-DocOwl2.}
\begin{tabular}{lc}
\toprule
\textbf{Rotation Angle} & \textbf{Avg. PDR (\%)} \\
\midrule
$0^\circ$               & 14.2 \\
$45^\circ$ (diagonal)   & 14.5 \\
$90^\circ$ (vertical)   & 14.1 \\
\bottomrule
\end{tabular}
\label{tab:rotation}
\end{table}

\subsection{Attention Heatmap Analysis Across \vlm}
\begin{figure}[h]
    \centering
    \includegraphics[width=1\linewidth]{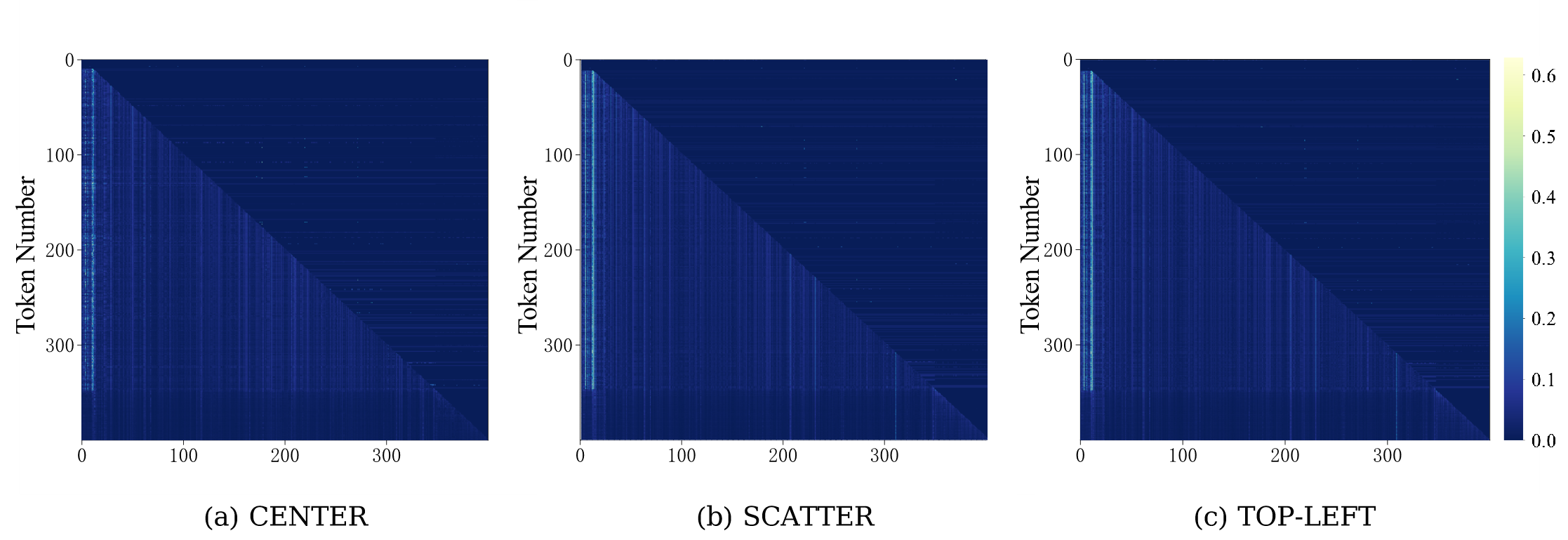}
    \caption{\textbf{Heatmap of attention wight variation} between watermarked multimodal input sequences and original one for \qwen. 
    }
    \label{fig:attention-weight-qwen}
\end{figure}

\begin{figure}[h]
    \centering
    \includegraphics[width=1\linewidth]{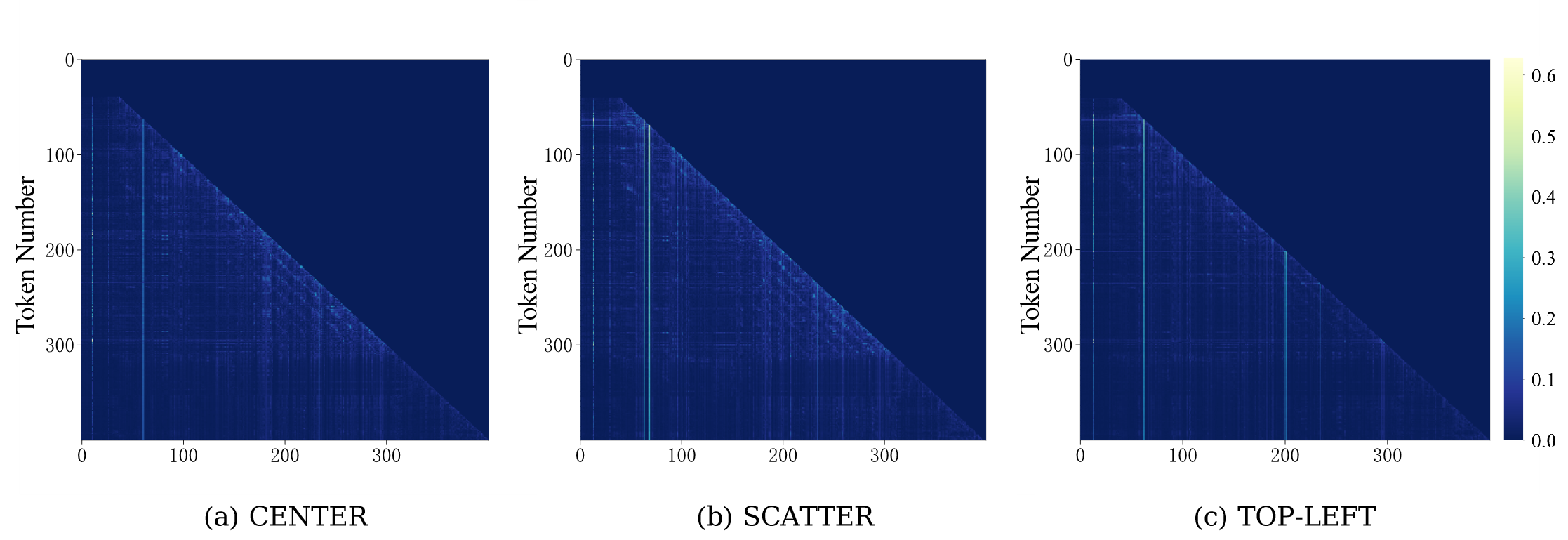}
    \caption{\textbf{Heatmap of attention wight variation} between watermarked multimodal input sequences and original one for \llava. 
    }
    \label{fig:attention-weight-llava}
\end{figure}

As shown in the \Cref{fig:attention-weight-qwen} and \Cref{fig:attention-weight-llava}, these are the attention variation heatmaps of \qwen and \llava. It is evident that the \qwen exhibits stronger robustness compared to \llava and \docowl, which is consistent with the experimental results.

\subsection{Image Preprocessing for Robustness Improvement}
\begin{figure*}[h]
    \centering
    \includegraphics[width=1\linewidth]{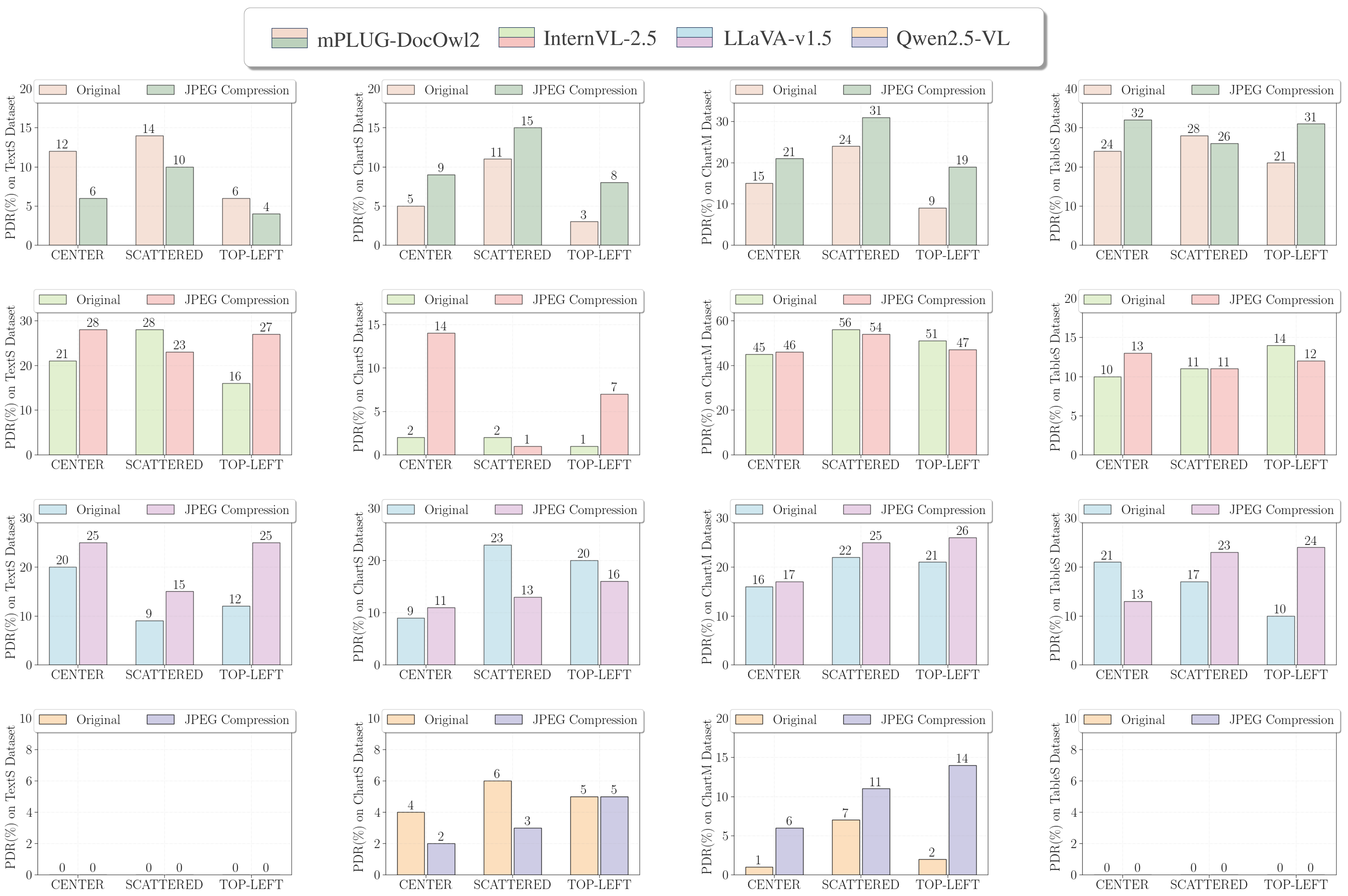}
    \caption{PDR before and after JPEG compression.}
    \label{fig:jpegcompression}
\end{figure*}

We try to preprocess document images before entering them into \vlm to reduce the interference effect of watermarking on the model. We test a common image processing method, JPEG compression, set the compression mass to 30, and analyze its effect on watermark interference. The experimental results are shown in \Cref{fig:jpegcompression}.

The experimental results show that the PDR value on the document data after JPEG compression is reduced compared to the unprocessed data in some cases.
The main mechanism of JPEG compression is to reduce the resolution of the watermark and thus reduce its perturbation effect. 
However, JPEG compression is essentially a lossy compression, which not only introduces additional noise, but also reduces the overall image quality, especially affecting the sharpness of the text area. 
This decrease in text resolution may affect \vlm' accurate extraction and understanding of document content, resulting in an increase on PDR. 

\subsection{Processing Architecture of \vlm}

\begin{figure*}[h]
    \centering
    \includegraphics[width=1\linewidth]{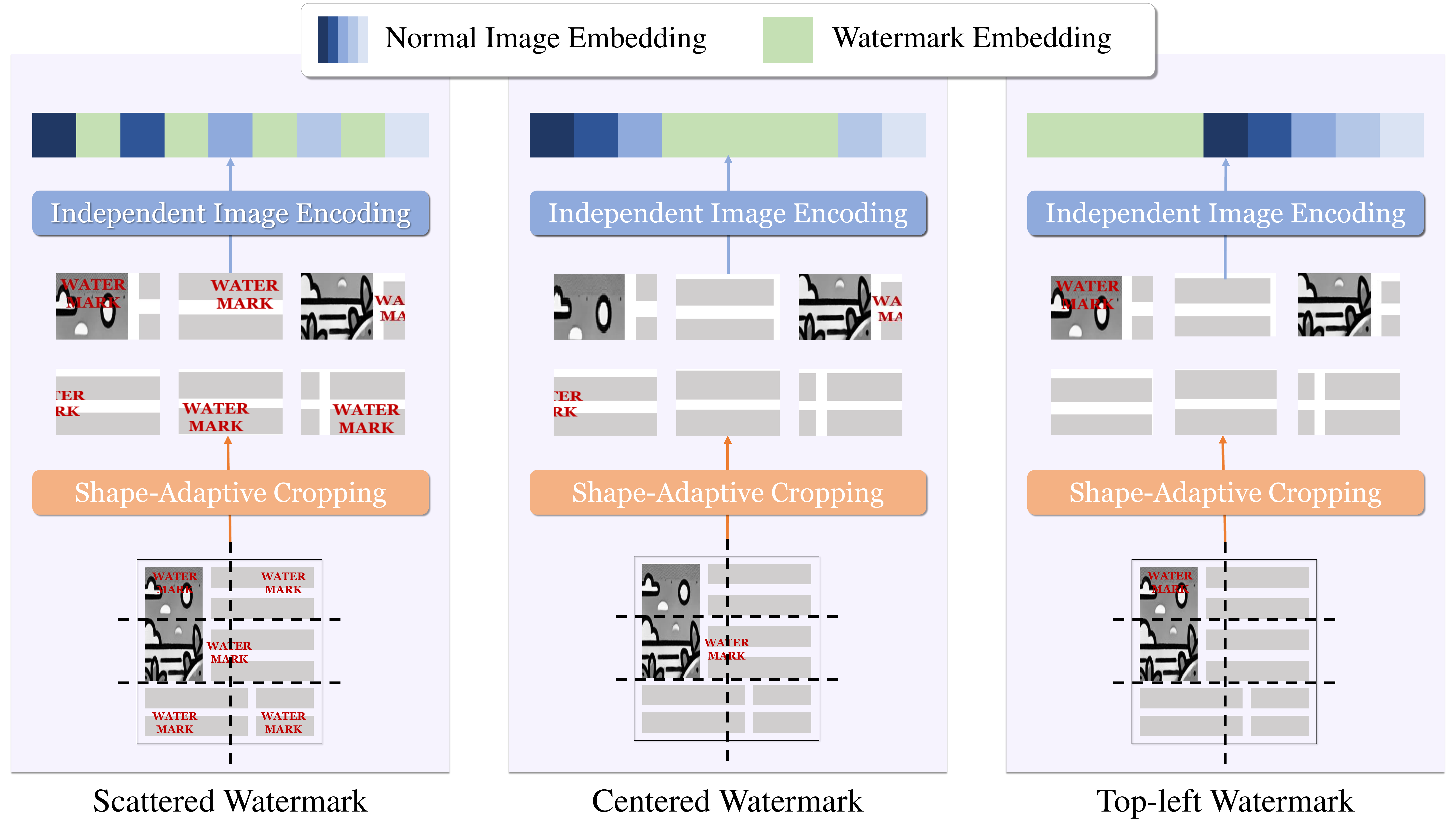}
    \caption{The structure of \vlm for processing input sequences.}
    \label{fig:docowl-framework}
\end{figure*}

As show in \Cref{fig:docowl-framework}, before performing multimodal fusion, the model first processes the input image by dividing it into multiple smaller patches. 
This step allows the model to analyze local features within each patch, facilitating more effective feature extraction and representation.

\end{document}